\documentclass[preprint,12pt]{elsarticle}

\usepackage{amssymb}
\usepackage{amsmath}

\begin{document}

\begin{frontmatter}



\title{Probabilistic Neural Networks (PNNs) for Modeling Aleatoric Uncertainty in Scientific Machine Learning}


\author[inst1]{Farhad Pourkamali-Anaraki}

\affiliation[inst1]{organization={Mathematical and Statistical Sciences},
            addressline={1201 Larimer St}, 
            city={Denver},
            postcode={80204}, 
            state={CO},
            country={USA}}

\author[inst2]{Jamal F. Husseini}
\author[inst2]{Scott E. Stapleton}

\affiliation[inst2]{organization={Mechanical and Industrial Engineering},
            addressline={1 University Ave}, 
            city={Lowell},
            postcode={01854}, 
            state={MA},
            country={USA}}

\begin{abstract}
This paper investigates the use of probabilistic neural networks (PNNs) to model aleatoric uncertainty, which refers to the inherent variability in the input-output relationships of a system, often characterized by unequal variance or heteroscedasticity. Unlike traditional neural networks that produce deterministic outputs, PNNs generate probability distributions for the target variable, allowing the determination of both predicted means and intervals in regression scenarios. Contributions of this paper include the development of a probabilistic distance metric to optimize PNN architecture, and the deployment of PNNs in controlled data sets as well as a practical material science case involving fiber-reinforced composites. The findings confirm that PNNs effectively model aleatoric uncertainty, proving to be more appropriate than the commonly employed Gaussian process regression for this purpose. Specifically, in a real-world scientific machine learning context, PNNs yield remarkably accurate output mean estimates with R-squared
  scores approaching 0.97, and their predicted intervals exhibit a high correlation coefficient of nearly 0.80, closely matching observed data intervals. Hence, this research contributes to the ongoing exploration of leveraging the sophisticated representational capacity of neural networks to delineate complex input-output relationships in scientific problems.
\end{abstract}


\end{frontmatter}


\section{Introduction}\label{sec:into}
Scientific data, ranging from observations and experiments to advanced computer simulations, enable the creation of predictive models, acting as digital twins of complex systems for the accurate prediction of behavior and optimal design \cite{nemani2023uncertainty}. The utility of machine learning surrogate models becomes particularly pronounced when it is challenging to derive a closed analytical formula linking input parameters directly with the desired outputs. Therefore, machine learning approaches have found application in diverse fields, such as material science \cite{johnson2020invited,nasrinpolymer,xie2023toward}, structural engineering \cite{pourkamali2021neural,tapeh2023artificial,thai2022machine}, environmental science \cite{weyn2020improving,salcedo2020machine}, and healthcare \cite{yang2019concepts,rajpurkar2022ai,daidone2024machine}, to name a few.  

In supervised machine learning, especially in regression tasks, training data sets are composed of input-output pairs. Typically, these data sets adhere to a deterministic model, meaning that a given input consistently produces the same output each time. However, this deterministic perspective ignores the inherent randomness or \textit{aleatoric uncertainty} present in complex real-world systems \cite{bae2021estimating,hullermeier2021aleatoric,zhang2023risk}. For example, in material science, properties such as tensile strength and electrical conductivity exhibit inherent variability due to variable processing conditions, material inconsistencies, and changes in environmental conditions. 
Even with advanced machinery, controlling every aspect of the manufacturing process is unrealistic, contributing to variability in these properties. Therefore, this randomness in scientific domains leads to a scenario in which a fixed input can generate a range of outputs, posing a challenge to the traditional deterministic approach of mapping inputs to outputs.

Consequently, it is crucial to develop machine learning models for surrogate modeling that accurately reflect the inherent variability or uncertainty of real-world systems. Nevertheless, most current machine learning methods fall short in providing reliable estimates of aleatoric uncertainty. For example, standard neural network regression models, such as multilayer perceptrons, typically provide point predictions because they employ deterministic transformations of the input data \cite{20-AOS1965,deshpande2022probabilistic,harakeh2023estimating}. Additionally, from a statistical viewpoint, the frequently employed mean squared error (MSE) loss function is based on the restrictive assumption that the output variance is constant and known for all inputs \cite{murphy2022probabilistic}. However, in reality, aleatoric uncertainty often exhibits heteroscedasticity \cite{yang2023explainable,immer2023effective}, meaning that variability is dependent on input. Therefore, practitioners and researchers involved in scientific machine learning face challenges at multiple stages, ranging from the development to the deployment of surrogate models.

To address these challenges, this paper examines the efficacy of probabilistic neural networks (PNNs) to model and quantify aleatoric uncertainty within scientific problems. PNNs, unlike standard neural networks, have the potential to go beyond point predictions by quantifying the inherent uncertainty of the data. The basic idea is to transform the final or output layer into a trainable probabilistic distribution, like a normal distribution \cite{maulik2020probabilistic,seitzer2022pitfalls}. This allows the mean and variance of this distribution to become parametric functions of inputs, enabling continuous weight updates guided by minimizing negative log likelihood. Consequently, PNNs adapt their predictive output distributions and uncertainty estimates, a crucial advantage over standard deterministic models blind to heteroscedastic aleatoric uncertainty. This enhanced versatility makes PNNs a powerful tool for surrogate modeling in scientific machine learning \cite{thiyagalingam2022scientific,psaros2023uncertainty}.

To effectively leverage PNNs for modeling aleatoric uncertainty, this paper details four pivotal contributions:
\begin{enumerate}
    \item \textit{Optimizing PNN architecture:} A key aspect of PNN development is determining the optimal architecture, which includes the depth (number of hidden layers) and the width (number of units per layer). In this paper, we introduce the use of Kullback-Leibler (KL) divergence \cite{ji2020kullback} to gauge the probabilistic distance between the actual and predicted output distributions. The proposed approach replaces traditional deterministic scoring metrics, such as the R-squared,  to optimize critical hyperparameters of PNNs, offering a more nuanced evaluation of model performance.
    \item \textit{Comparative analysis with other models:} We conduct a comprehensive comparison between PNNs and other machine learning models primarily designed to capture model or epistemic uncertainty \cite{hullermeier2021aleatoric}. Our findings demonstrate that popular models like Gaussian process regression \cite{karvonen2023maximum}, although producing a normal distribution for the  output variable, fail to accurately estimate aleatoric uncertainty, even in synthetic and controlled data settings. This underscores the significance of selecting appropriate models to model aleatoric uncertainty.
    \item \textit{Benchmarking with the Ishigami function:} We apply PNNs to a well-known benchmark in scientific machine learning and uncertainty quantification: the Ishigami function \cite{hariri2022structural}. This function, characterized by its high nonlinearity and heteroscedastic noise, serves as an excellent test case to illustrate the effectiveness of PNNs in complex scenarios.
    \item \textit{Real-world Application in Material Science:} We investigate a practical scientific machine learning problem involving the development of surrogate models for fiber-reinforced composite microstructure generators. Given the randomness in initial seeding and the amplification of variability through the ensuing discrete element simulations, it is crucial to quantify the heteroscedastic aleatoric uncertainty for various input conditions. 
\end{enumerate}

The remainder of this paper is organized as follows. In Section \ref{sec:prior}, we dive into the foundations of deterministic neural network models, focusing on data transformation methods and the MSE loss function, which originates from the maximum likelihood estimation framework. Section \ref{sec:proposed} is dedicated to explaining the core components of PNNs and the adaptation of the loss function to incorporate heteroscedastic aleatoric uncertainty using the negative log likelihood principle. Additionally, we introduce a novel evaluation metric tailored for neural architecture search that accounts for the probabilistic nature of outputs. We evaluate the performance of PNNs using the Ishigami function in Section \ref{sec:exp-syn}, a standard benchmark for testing machine learning surrogate models in complex nonlinear scenarios. Section \ref{sec:exp-real} discusses the application of PNNs to a large-scale and real-world problem in materials science, specifically the computational modeling of fiber-reinforced composites. To conclude this paper, Section \ref{sec:conc} offers a summary of our findings, practical implications, and directions for future research, particularly in the context of distinguishing various sources of uncertainty in neural network regression models.

\section{Overview of Deterministic Neural Networks}\label{sec:prior}
A key element in training machine learning models involves minimizing a carefully chosen loss function. This minimization is essential to narrow the gap between actual and predicted outputs. By conceptualizing training as an estimation problem, we can employ a statistical model that links inputs with their corresponding outputs. This approach transforms the training process into the maximum likelihood estimation (MLE) problem. To clarify this concept, we assume that the conditional probability of the output $y\in\mathbb{R}$, given the input vector $\mathbf{x}\in\mathbb{R}^D$, adheres to a normal distribution:
\begin{equation}
p(y|\mathbf{x};\boldsymbol{\theta})=\mathcal{N}\big(f_{\mu}(\mathbf{x};\boldsymbol{\theta}), f_{\sigma}(\mathbf{x};\boldsymbol{\theta})\big),\label{eq:cond-prob}
\end{equation}
where $\boldsymbol{\theta}$ denotes all unknown parameters that we aim to infer during the training process. Furthermore, $f_{\mu}(\mathbf{x};\boldsymbol{\theta})\in\mathbb{R}$ and $f_{\sigma}(\mathbf{x};\boldsymbol{\theta})\in\mathbb{R}_{+}$ take the place of the mean and variance of the normal distribution, respectively. For simplicity, it is common to assume that the variance is fixed. Thus, the function $f_{\sigma}(\mathbf{x};\boldsymbol{\theta})$ is replaced by a known scalar $\sigma^2$,  which is independent of the input vector $\mathbf{x}$. As a result, we can rewrite the conditional distribution in Eq.~\eqref{eq:cond-prob} as a deterministic function of the input plus a zero-mean fixed-variance normal distribution, i.e., $p(y|\mathbf{x};\boldsymbol{\theta})=f_{\mu}(\mathbf{x};\boldsymbol{\theta}) + \varepsilon, \varepsilon\sim \mathcal{N}(0, \sigma^2)$, thereby assuming that the aleatoric or data uncertainty is homoscedastic. 

With this assumption in place, the objective of MLE is to determine the model parameters that maximize the likelihood across all input-output pairs in the training data set. This is typically achieved by maximizing the logarithm of the  likelihood function, or equivalently minimizing negative log likelihood (NLL), for a given training data set in the form of $\mathcal{D}_{\text{train}}=\{(\mathbf{x}_i, y_i)\}_{i=1}^n$:
\begin{align}
    \text{NLL}(\boldsymbol{\theta})&=-\sum_{i=1}^n \log p(y_i|\mathbf{x}_i;\boldsymbol{\theta})\nonumber \\
    &=-\sum_{i=1}^n \log \Big[\Big(\frac{1}{2\pi \sigma^2}\Big)^{1/2} \exp\Big(-\frac{1}{2\sigma^2} \big(y_i - f_{\mu}(\mathbf{x}_i;\boldsymbol{\theta})\big)^2\Big) \Big]\nonumber \\
&=  \frac{n}{2}\log(2\pi \sigma^2) + \frac{1}{2\sigma^2}\sum_{i=1}^n \big(y_i - f_{\mu}(\mathbf{x}_i;\boldsymbol{\theta})\big)^2.\label{eq:nll}
\end{align}
Therefore, in the case where the variance $\sigma^2$ is fixed, the first term $\frac{n}{2}\log(2\pi \sigma^2)$ becomes a constant with respect to $\boldsymbol{\theta}$ and is thus irrelevant for optimization. As a result, if NLL in Eq.~\eqref{eq:nll}  is multiplied by $\frac{2\sigma^2}{n}$, it simplifies to the widely used mean squared error (MSE) loss function.

In summary, this simplification process used to derive the MSE loss function results in a deterministic model in the form of $\mathbf{x}\mapsto f_{\mu}(\mathbf{x};\boldsymbol{\theta}^*)$. In this representation, $\boldsymbol{\theta}^*$
  signifies the parameters that minimize the NLL function in Eq.~\eqref{eq:nll}. This deterministic model, particularly in the context of neural networks, often involves a complex choice of parameterization. For example, multilayer perceptrons (MLPs), also known as fully connected feed-forward networks, are characterized by their layered architecture \cite{liu2017survey}. These layers collectively form a composite function where the output of one layer serves as the input to the next, leading to the following mapping:
  \begin{align}
    \text{input layer: }&\mathbf{x}^{(0)}=\mathbf{x}, \nonumber \\
    \text{hidden layers: }&\mathbf{x}^{(l)}= g^{(l)}\big(\mathbf{W}^{(l)}\mathbf{x}^{(l-1)}+\mathbf{b}^{(l)}\big),\;l=1,\ldots,L, \nonumber \\
    \text{output layer: }&f_{\mu}(\mathbf{x};\boldsymbol{\theta}) = g^{(L+1)}\big(\mathbf{W}^{(L+1)}\mathbf{x}^{(L)}+\mathbf{b}^{(L+1)}\big).\label{eq:nn}
\end{align}
Therefore, the prediction model $f_{\mu}(\mathbf{x};\boldsymbol{\theta})$ takes on a composite or nested form, where $\boldsymbol{\theta}=\{\mathbf{W}^{(1)}, \ldots,\mathbf{W}^{(L+1)}, \mathbf{b}^{(1)}, \ldots,\mathbf{b}^{(L+1)}\}$ contains the weight matrix and the bias vector for each layer. Therefore, the processing of the input vector $\mathbf{x}$ in standard neural networks is deterministic, meaning that every time the same input is provided, the network will produce the exact same predicted output. However, this deterministic behavior limits the ability of neural networks to naturally provide estimates of aleatoric uncertainty in scientific applications. In Section \ref{sec:proposed}, we will explain the building blocks of probabilistic neural networks to provide uncertainty estimates. 

Beyond the challenges of producing predictive uncertainties, inferring the model parameters $\boldsymbol{\theta}$ during training requires the selection of various hyperparameters in advance.  For example, the number of hidden layers $L$ and the number of units or neurons in each hidden layer are pivotal hyperparameters that determine the depth and width of the network, respectively \cite{POURKAMALIANARAKI2023106983}. To identify a neural network configuration that is sufficiently expressive for the specific problem being addressed, it is often essential to experiment with different combinations of network depth and width \cite{chitty2023neural,jin2023autokeras}. 
Another hyperparameter includes the selection of nonlinear activation functions, $g^{(l)}$, such as the Rectified Linear Unit (ReLU), which sets negative values to zero, and the Exponential Linear Unit (ELU) that allows a smoother treatment of negative values \cite{jagtap2023important}. For the purposes of this study, we adopt the ELU activation function, which is defined as follows:  
\begin{equation}
\text{ELU}(z)=\begin{cases}
    z & \text{if } z > 0 \\
 e^z - 1 & \text{if } z \leq 0
\end{cases}.
\end{equation}

In the final part of this section, we explain the evaluation and scoring techniques used to assess the accuracy of regression models. Typically, for a given test set $\mathcal{D}_{\text{test}}=\{(\mathbf{x}_i^{(t)}, y_i^{(t)})\}_{i=1}^{n_t}$, we take a set of predicted and actual outputs to compute a measure of the deviation between them. Although the MSE loss function can be used for this purpose during the testing phase, the R-squared score or the coefficient of determination emerges as a preferred alternative in many problems. The R-squared score is defined by the following equation \cite{renaud2010robust}:
\begin{equation}
    R^2(y_{\text{test}}, \hat{y}_{\text{test}}) = 1 - \frac{\sum_{i=1}^{n_t} \big(y_i^{(t)} - \hat{y}_i^{(t)}\big)^2}{\sum_{i=1}^{n_t} \big(y_i^{(t)} - \bar{y}^{(t)}\big)^2},
\end{equation}
where $\hat{y}_i^{(t)}$ is the predicted value for the $i$-th testing sample and $y_i^{(t)}$ is the corresponding true output, for each $i$ ranging from $1$ to $n_t$. Thus, $n_t$ represents the total number of testing samples and $\bar{y}^{(t)} = \frac{1}{n} \sum_{i=1}^{n} y_i^{(t)}$ denotes the mean of the actual outputs. 

An R-squared score of 1 signifies that the model's predictions perfectly match the actual data, indicating excellent predictive accuracy. Conversely, an R-squared
  score near 0 implies that the model's predictions do not adequately reflect the variance within the set $\mathcal{D}_{\text{test}}$, essentially suggesting that the model's predictions are no better than simply guessing the mean value $\bar{y}^{(t)}$
  for all predictions. Thus, the 
R-squared score, which can reach a maximum value of 1, provides a normalized measure of the regression model's predictive performance.
However, it is important to note a significant limitation of the 
R-squared score: it is designed for evaluating deterministic models and is not suitable for assessing models that make probabilistic predictions to measure the confidence level. Therefore, in the next section, we will introduce an enhanced evaluation metric designed to assess similarities between actual and predicted probability distributions. 

\section{Building Blocks of Probabilistic Neural Networks (PNNs)}\label{sec:proposed}
PNNs represent a class of artificial neural networks that integrate probability distributions within their multilayered transformations, discussed in Eq.~\eqref{eq:nn}. These networks are particularly effective in capturing the intricate relationships between inputs and outputs within a probabilistic framework, especially in scenarios characterized by data or aleatoric uncertainty. This section begins with developing a suitable loss function for scientific problems encompassing heteroscedastic aleatoric uncertainty, utilizing the MLE framework without the fixed-variance assumption. Following this, we introduce the concept of incorporating a probabilistic layer as the output layer in a network. Additionally, we describe a method for evaluation purposes that is cognizant of uncertainty, aiding in the determination of the optimal network depth and width. We also present a controlled benchmark problem to facilitate the comparison of PNNs with Gaussian process regression (GPR), a technique extensively employed in prior research for uncertainty quantification.

\subsection{Designing Loss Functions for Aleatoric Uncertainty Quantification}
In this section, we revisit the statistical model for the output variable, which takes the form of a conditional probability as shown in Eq.~\eqref{eq:cond-prob}. The goal is to eliminate the restrictive assumption that the variance term, that is, $f_{\sigma}(\mathbf{x};\boldsymbol{\theta})$, must be fixed and known in advance. To this end, we compute the negative log likelihood or NLL function using both the mean and variance functions: 
\begin{align}
    \text{NLL}(\boldsymbol{\theta})&=-\sum_{i=1}^n \log p(y_i|\mathbf{x}_i;\boldsymbol{\theta})\nonumber \\
    &=-\sum_{i=1}^n \log \Big[\Big(\frac{1}{2\pi f_{\sigma}(\mathbf{x}_i;\boldsymbol{\theta})}\Big)^{1/2} \exp\Big(-\frac{\big(y_i - f_{\mu}(\mathbf{x}_i;\boldsymbol{\theta})\big)^2}{2f_{\sigma}(\mathbf{x}_i;\boldsymbol{\theta})} \Big) \Big]\nonumber\\
    & = \frac{1}{2} \sum_{i=1}^n \log \Big(2\pi f_{\sigma}(\mathbf{x}_i;\boldsymbol{\theta}) \Big) + \frac{1}{2}\sum_{i=1}^n \frac{\big(y_i - f_{\mu}(\mathbf{x}_i;\boldsymbol{\theta})\big)^2}{f_{\sigma}(\mathbf{x}_i;\boldsymbol{\theta})}.\label{eq:nll-full}
\end{align}
The NLL function consists of two parts: the first term involves the logarithm of the variance function $f_{\sigma}(\mathbf{x}_i;\boldsymbol{\theta})$, which accounts for the uncertainty of the prediction. The second term is a scaled squared difference between the actual output value $y_i$ and the predicted mean $f_{\mu}(\mathbf{x}_i;\boldsymbol{\theta})$, which penalizes deviations of the model's predictions from the true outputs. The scaling factor is the inverse of the predicted variance, emphasizing the model's confidence in its predictions. To further explain the scaling factor, note that when the model is certain $f_{\sigma}(\mathbf{x}_i;\boldsymbol{\theta})$ is small. Therefore, the inverse scaling factor 
becomes larger. Now, the penalty for the difference between the predicted mean and the actual value is increased. Hence, the model is confident about its prediction, so if it makes an inaccurate prediction, it is penalized more heavily. 

Minimizing the NLL function with respect to $\boldsymbol{\theta}$ during the training stage encourages the model to accurately predict the mean of the target variable while also estimating the corresponding variance. The derivatives of the NLL function with respect to the predicted mean $f_{\mu}(\mathbf{x};\boldsymbol{\theta})$
and the predicted variance 
$f_{\sigma}(\mathbf{x};\boldsymbol{\theta})$ are pivotal for gradient-based optimization methods. These techniques are integral to deep learning frameworks, facilitating the step-by-step refinement of the model parameters 
$\boldsymbol{\theta}$. To ensure a comprehensive understanding, we list the derivatives of the NLL function with respect to both the predicted mean and variance. This elucidates the mechanism of parameter updating during model training. For the sake of simplicity and to make the explanation more straightforward, we reduce the NLL loss function to a scenario involving only a single input-output pair $(\mathbf{x}, y)$ \cite{stirn2023faithful}:
\begin{align}
    &\frac{\partial \text{NLL}}{\partial f_{\mu}(\mathbf{x};\boldsymbol{\theta})}=-\frac{(y - f_{\mu}(\mathbf{x};\boldsymbol{\theta}))}{f_{\sigma}(\mathbf{x};\boldsymbol{\theta})}, \nonumber \\
    & \frac{\partial \text{NLL}}{\partial f_{\sigma}(\mathbf{x};\boldsymbol{\theta})}=\frac{f_{\sigma}(\mathbf{x};\boldsymbol{\theta}) - (y - f_{\mu}(\mathbf{x};\boldsymbol{\theta}))^2}{2(f_{\sigma}(\mathbf{x};\boldsymbol{\theta}))^2}.\label{eq:grad}
\end{align}
Therefore, when the model's predicted variance $f_{\sigma}(\mathbf{x};\boldsymbol{\theta})$
  is small, indicating high confidence, the gradients regarding both $f_{\mu}(\mathbf{x};\boldsymbol{\theta})$ and $f_{\sigma}(\mathbf{x};\boldsymbol{\theta})$
  lead to more substantial updates. This mechanism ensures that the model's adjustments are more pronounced when it is confident in its predictions, demanding accuracy in such cases, while allowing for more cautious updates when the model acknowledges greater uncertainty.
  
\subsection{Trainable Distributions in Probabilistic Layers}
\begin{figure}[ht]
    \centering
    \includegraphics[width=\textwidth]{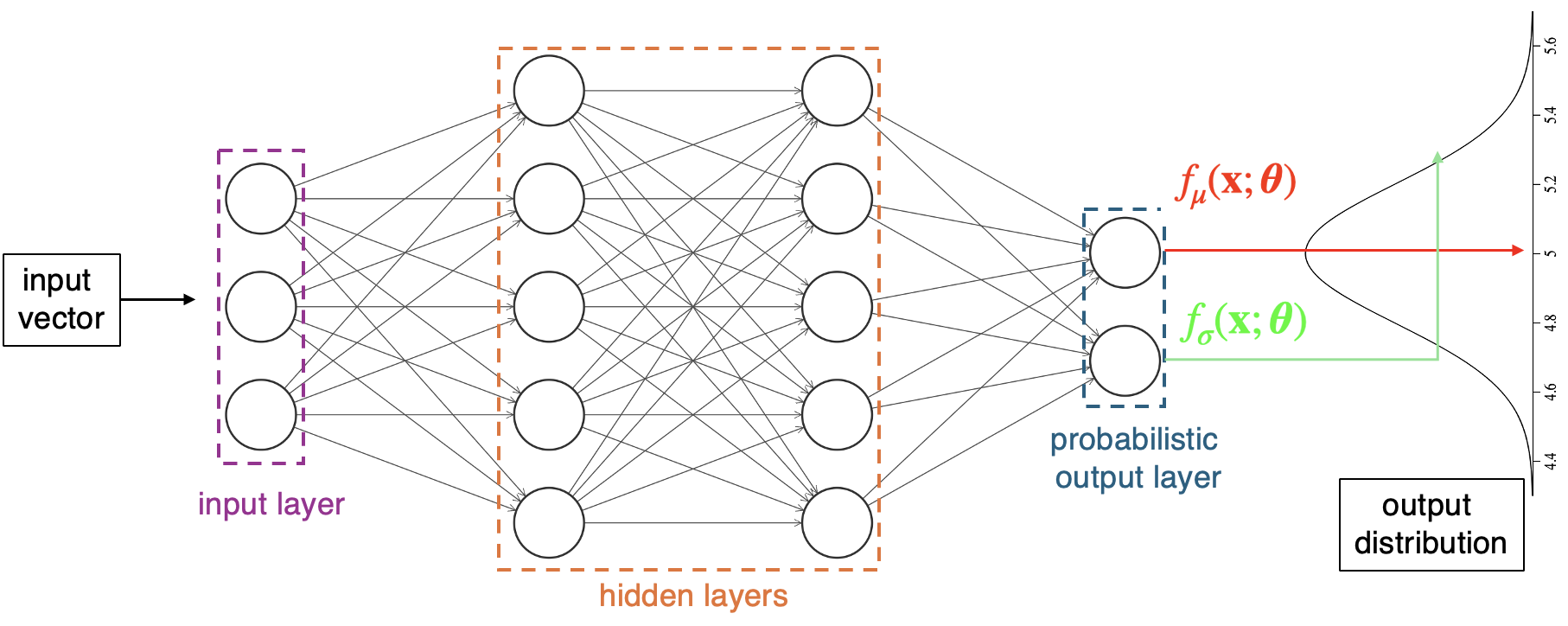}
    \caption{The inclusion of the probabilistic output layer transforms the neural network from making deterministic predictions to providing a normal distribution characterized by its mean $f_{\mu}(\mathbf{x};\boldsymbol{\theta})$ and variance $f_{\sigma}(\mathbf{x};\boldsymbol{\theta})$. By providing a distribution of possible outcomes rather than a single value, PNNs enable the quantification of heteroscedastic aleatoric uncertainty.}
    \label{fig:pnn}
\end{figure}

Incorporating a probabilistic layer in neural networks marks a pivotal shift, enabling these networks to produce more than mere point predictions; they generate distributions representing a range of possible outcomes. This innovative layer is adeptly engineered to yield two critical outputs for each input vector $\mathbf{x}$: the mean, denoted as 
$f_{\mu}(\mathbf{x};\boldsymbol{\theta})$, and the variance, represented by 
$f_{\sigma}(\mathbf{x};\boldsymbol{\theta})$. Consequently, PNNs exhibit a dual-headed architecture, with one ``head'' focused on predicting the expected value of the target variable and the other dedicated to quantifying the uncertainty associated with this prediction. Figure \ref{fig:pnn} visually illustrates this unique architecture of PNNs with the dual-headed structure. 

Given that the variance in a normal distribution inherently signifies the data's spread or variability, it must always remain positive. To enforce this property within PNNs, the Softplus function is often employed \cite{zheng2015improving,nag2023serf}. This function effectively maps any real number 
$z$ to a positive domain using the transformation 
$\log(1+\exp(z))$. The exponential component guarantees that the output remains positive, and by adding 1, the expression 
$1+\exp(z)$ is assured to be strictly greater than 1. The subsequent logarithmic operation then smoothly scales the output, ensuring that it is positive and continuously differentiable, a crucial aspect for maintaining the mathematical and computational integrity of the model.

Therefore, incorporating trainable distributions into the final layer of PNNs enables the identification of model parameters, denoted as $\boldsymbol{\theta}$, which include layer weights and biases, maximizing the likelihood of generating the observed input-output data pairs. This approach diverges from traditional methods that focus solely on minimizing the discrepancy between actual outputs and point predictions, such as the MSE loss function. Instead, PNNs employ a more rigorous strategy that accounts for data uncertainty or variability by optimizing both the predicted mean $f_{\mu}(\mathbf{x};\boldsymbol{\theta})$ and variance $f_{\sigma}(\mathbf{x};\boldsymbol{\theta})$ of the output distribution. The training process is facilitated by minimizing the NLL loss function in Eq.~\eqref{eq:nll-full}, which is dependent on the predicted mean and variance. The closed-form expressions for the derivatives of the NLL with respect to these two variables, provided in Eq.~\eqref{eq:grad}, 
  are critical for the application of gradient-based optimization techniques. These gradients enable the systematic adjustment of model parameters
$\boldsymbol{\theta}$ through backpropagation.

In this paper, our primary focus is on the Root Mean Square Propagation (RMSProp) optimization algorithm \cite{mukkamala2017variants,abdulkadirov2023survey}. RMSProp dynamically adjusts the learning rates for each parameter by computing a moving average of the squared gradients. This approach effectively mitigates the issue of updates becoming negligible due to small gradients (vanishing updates) and overly large due to significant gradients (exploding updates). Formally, let  $\mathbf{g}:=\nabla_{\boldsymbol{\theta}}\text{NLL}$
represent the gradient of the NLL loss function with respect to the model parameters. The update mechanism for the model parameters is then described by the following equations:
\begin{align}
    \mathbf{s}&\leftarrow\gamma \;\mathbf{s} + (1-\gamma)\; \mathbf{g}^2, \nonumber\\
\boldsymbol{\theta} & \leftarrow \boldsymbol{\theta} - \frac{\eta}{\sqrt{\mathbf{s}+\epsilon}}\mathbf{g}.
\end{align}
In this context, $\mathbf{s}$ refers to the updated moving average of squared gradients. The decay rate, denoted by $\gamma$, is typically chosen as $0.9$. The learning rate, represented by $\eta$, is fixed at $0.001$ in this study. Additionally, $\epsilon$ is a small constant incorporated to enhance numerical stability.
 
Therefore, this update mechanism can be used to find the layer weights and biases in PNNs, which we refer to them as $\boldsymbol{\theta}^*$. Upon the completion of the training phase, PNNs leverage these optimized parameters to predict the mean and variance of the output distribution for new input test points, denoted by $\mathbf{x}^{(t)}$. This prediction is facilitated through two distinct function evaluations: $f_{\mu}(\mathbf{x}^{(t)}; \boldsymbol{\theta}^*)$ for the mean and $f_{\sigma}(\mathbf{x}^{(t)}; \boldsymbol{\theta}^*)$ for the variance, providing a comprehensive statistical analysis of the network's output.

It is important to note that the primary focus of this paper is on modeling aleatoric uncertainty, a crucial but often neglected aspect in scientific machine learning. Consequently, our discussion is limited to the integration of probabilistic layers at the network's final stage; see Figure \ref{fig:pnn}. While it is feasible to introduce probability distributions to the weights and biases across all network layers, leading to what is commonly known as Bayesian neural networks (BNNs) \cite{magris2023bayesian,oleksiienko2023variational}, such considerations fall outside the purview of this paper.

\subsection{Optimizing the Architecture of PNNs}

Neural networks derive their remarkable expressive capabilities from the incorporation of hidden layers and units \cite{raghu2017expressive,lu2021deep,adcock2021gap}; see the intermediate layers in Figure \ref{fig:pnn}. As such, a pivotal aspect of neural network development involves determining the optimal depth (number of hidden layers) and width (number of units or neurons per layer). While grid search remains a prevalent method for fine-tuning neural network architectures, the success of this optimization process lies in the selection of a suitable evaluation or scoring metric. Such metrics are essential for effectively comparing the performance of various models, ensuring that the chosen architecture not only fits the data well but also generalizes effectively to new, unseen data. This balance between model complexity and performance is crucial for the development of robust and efficient neural networks.

As highlighted in Section \ref{sec:prior}, prevalent evaluation metrics like the R-squared score predominantly focus on the mean prediction $f_{\mu}(\mathbf{x};\boldsymbol{\theta})$ and often overlook variance-related information. This oversight is particularly critical in the context of neural network regression models tailored for scientific applications, where aleatoric or inherent data uncertainty plays a significant role. To address this gap, this work introduces an evaluation metric in the context of PNNs that takes into account the entire distribution of actual and predicted outputs, thereby offering a more comprehensive assessment that aligns with the complexities of scientific data.

To explain our methodology, we start by identifying \textit{unique} input vectors within the test data set $\mathcal{D}_{\text{test}}$. It is important to remember that due to aleatoric uncertainty, a single input vector can lead to a variety of output values. Consequently, for each distinct input vector in the test set, represented as $\mathbf{x}^{(t)}$, we calculate the empirical mean and variance of the associated outputs. This process allows us to approximate the distribution of the actual output variables. We model this distribution as a normal distribution, denoted by $\mathcal{N}(f_{\mu}^{\text{emp}}({\mathbf{x}^{(t)}}), f_{\sigma}^{\text{emp}}({\mathbf{x}^{(t)}}))$, where $f_{\mu}^{\text{emp}}({\mathbf{x}^{(t)}})$ and $f_{\sigma}^{\text{emp}}({\mathbf{x}^{(t)}})$ represent the mean and variance of the actual outputs for the input vector $\mathbf{x}^{(t)}$. Note that these two quantities do not depend on the optimized parameters of the trained PNN, i.e.,  $\boldsymbol{\theta}^*$. 
This approach provides a statistical basis for evaluating the predictive performance of our models, taking into account both the central tendency and the dispersion of the predicted values.

Next, we employ the Kullback-Leibler (KL) divergence \cite{feng2018learning,guo2024off} to measure the probabilistic distance between the actual and predicted output distributions. That is, we compute the predicted mean and variance for the input test data point according to the trained PNN, i.e., $f_{\mu}(\mathbf{x}^{(t)}; \boldsymbol{\theta}^*)$ and $f_{\sigma}(\mathbf{x}^{(t)}; \boldsymbol{\theta}^*)$, utilizing the model parameters $\boldsymbol{\theta}^*$ obtained from training. Given the two normal distributions, the KL divergence can be calculated analytically due to the well-defined properties of normal distributions:
\begin{align}
    &D_{KL}\Big(\mathcal{N}\big(f_{\mu}^{\text{emp}}({\mathbf{x}^{(t)}}), f_{\sigma}^{\text{emp}}({\mathbf{x}^{(t)}})\big) \;||  \;\mathcal{N}\big(f_{\mu}({\mathbf{x}^{(t)}};\boldsymbol{\theta}^*), f_{\sigma}({\mathbf{x}^{(t)}};\boldsymbol{\theta}^*)\big)\Big)\nonumber \\
    &=\frac{1}{2}\log\Big(\frac{f_{\sigma}({\mathbf{x}^{(t)}};\boldsymbol{\theta}^*)}{f_{\sigma}^{\text{emp}}({\mathbf{x}^{(t)}})}\Big) + \frac{f_{\sigma}^{\text{emp}}({\mathbf{x}^{(t)}}) + \big(f_{\mu}^{\text{emp}}({\mathbf{x}^{(t)}}) - f_{\mu}({\mathbf{x}^{(t)}};\boldsymbol{\theta}^*)\big)^2}{2 f_{\sigma}({\mathbf{x}^{(t)}};\boldsymbol{\theta}^*)} -\frac{1}{2}. \label{eq:KL}
\end{align}
Note that the KL divergence is zero when the actual and predicted distributions are identical, reflecting perfect agreement. The KL divergence is always non-negative, and it increases as the two distributions diverge further from each other, quantifying the amount of information lost when one distribution is used to approximate the other. Thus, in the context of PNNs, the KL divergence not only provides a more systematic and principled approach for evaluation compared to traditional deterministic methods used for grid searching various depth and width configurations, but it also offers a cost-effective advantage due to its closed-form solution when measuring the distance between two normal distributions.

\subsection{Controlled Environment Case Study: Dissecting Aleatoric and Epistemic Uncertainty}

In this section, we examine a 1D synthetic data set to showcase the capability of PNNs in modeling a certain type of heteroscedastic aleatoric noise. This case study, set in a controlled environment, allows us to explore the effectiveness of KL divergence for choosing the best model, as outlined in Eq.~\eqref{eq:KL}. We particularly emphasize the significance of two critical hyperparameters in model selection: the number of hidden layers (which determines the model's depth) and the number of units in each hidden layer (which defines the model's width). For simplicity, we assume a uniform width across all hidden layers. In this study, we employ TensorFlow/Keras for neural network training, and all experiments were performed on a MacBook Pro equipped with an Apple M2 Max chip and 32 GB of RAM. 

The initial step in this case study involves considering a cubic function that establishes the following relationship between scalar inputs $x$ and scalar outputs $y$:
\begin{equation}
    y = x^3 + \underbrace{0.1 (2+x)\varepsilon}_{\text{heteroscedastic noise}},\label{eq:data}
\end{equation}
where $x$ lies in the interval $[-1,1]$ and $\varepsilon$ is drawn from the standard normal distribution $\mathcal{N}(0,1)$. To construct the training data set $\mathcal{D}_{\text{train}}$, we use uniform sampling to select $100$ points from the interval $[-1,1]$. For each of these points, we generate $10$ instances of the random variable $\varepsilon$, leading to a heteroscedastic noise model. Consequently, the training data set comprises a total of $1,\!000$ input-output pairs. Adopting a similar approach for the test data set $\mathcal{D}_{\text{test}}$, we select $50$ points from the same interval $[-1,1]$ and generate $10$ realizations of $\varepsilon$ for each, resulting in a total of $500$ test input-output pairs and $50$ distinct input conditions for testing.

\begin{figure}[ht]
    \centering
    \includegraphics[width=\textwidth]{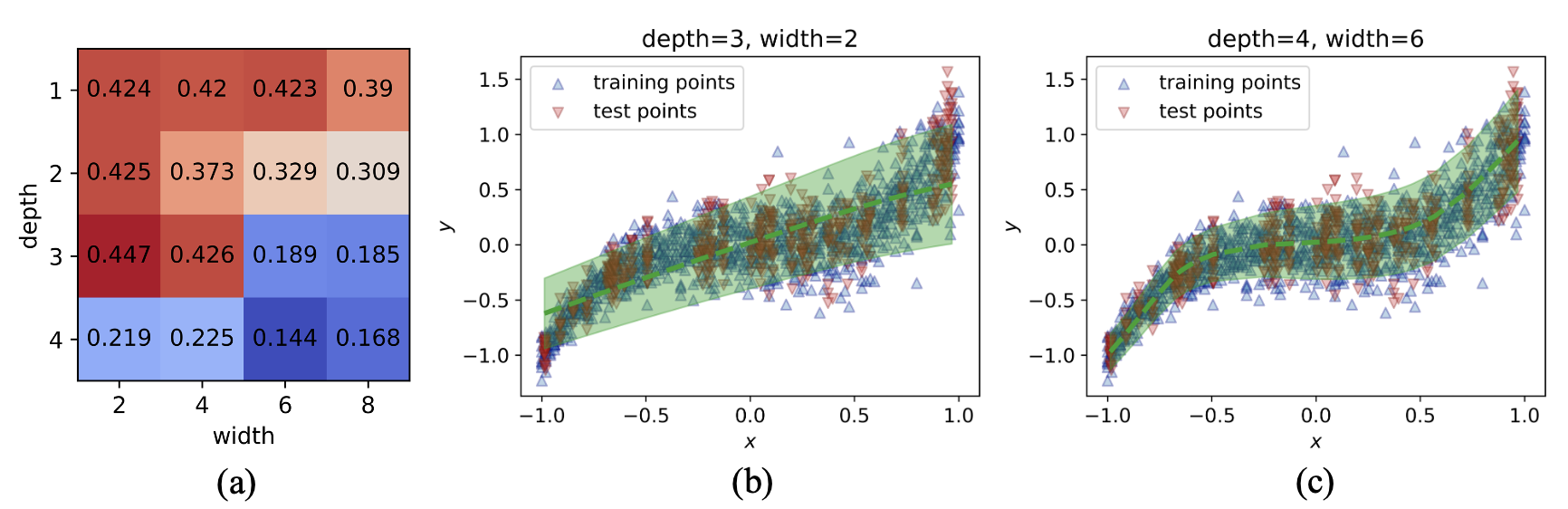}
    \caption{Demonstrating the critical role of hyperparameter tuning in PNNs through the application of the proposed KL divergence metric. This controlled case study explores $4$ varying depths for the number of hidden layers, ranging from $1$ to $4$, alongside $4$ distinct widths, represented by the number of hidden units within the set $\{2, 4, 6, 8\}$. The outcomes of the grid search are showcased in (a), while  (b) and (c) respectively highlight the least effective and optimal PNN models identified through this process. Note that in (a), lower KL divergence values are favored as they indicate a closer match between the actual and predicted output distributions.}
    \label{fig:syn_visual}
\end{figure}

Figure \ref{fig:syn_visual}(a) reports the results of a grid search carried out on PNNs that vary in depth and width. For this study, the batch size was maintained at $32$, and each model underwent training over $100$ epochs with the discussed RMSProp optimizer. The analysis, guided by the observed Kullback-Leibler (KL) divergences, reveals that a single hidden layer in PNNs fails to offer sufficient expressiveness, regardless of the width setting. Conversely, increasing the network's depth appears to bring the model's predicted output distribution more in alignment with the true distribution. It should be noted that a PNN configuration with $4$ hidden layers and $6$ units in each layer achieved the lowest KL divergence, marking it as the top-performing model. On the contrary, a model configured with $3$ hidden layers and a width of $2$ units exhibited the poorest performance, recording a KL divergence of $0.447$, which is nearly triple that of the best-performing model. Therefore, this case study underscores the critical role of model selection in PNN applications, highlighting that mere increases in depth or width do not guarantee improved predictive accuracy.

To further underscore the importance of selecting appropriate PNN architectures for modeling aleatoric uncertainty, we present visual comparisons of the worst and best predictive models in Figures \ref{fig:syn_visual}(b) and \ref{fig:syn_visual}(c), respectively. The dotted curve in these figures depicts the mean function $f_{\mu}$, akin to that in deterministic neural networks. Additionally, we highlight prediction intervals through shaded green areas, representing the lower and upper bounds of a 90\% confidence interval for a normal distribution, determined by the mean function $f_{\mu}$
  and the variance function $f_{\sigma}$. 
While the least accurate PNN manages to follow the general input-output relationship, it inadequately represents aleatoric uncertainty, particularly evident in the overly broad prediction intervals for smaller values of $x$. In contrast, the optimized PNN in  Figure \ref{fig:syn_visual}(c) markedly improves in quantifying aleatoric uncertainty, offering a more precise estimate. This comparison clearly demonstrates the potential of well-optimized PNNs to deliver accurate and reliable aleatoric uncertainty estimations. 

Next, we evaluate the performance of the best performing PNN, characterized by a depth of $4$ and width of $6$, by graphically comparing the predicted mean with the actual or empirical mean, alongside the predicted intervals against the actual intervals for each test point. The empirical intervals are determined by computing the range, which is the difference between the maximum and minimum of $10$ outputs for each input vector $\mathbf
{x}^{(t)}$ from the test set $\mathcal{D}_{\text{test}}$. A key advantage of this assessment methodology is its adaptability to more complex scenarios, making it applicable to regression problems in higher dimensional input spaces. 

\begin{figure}[ht]
    \centering
    \includegraphics[width=\textwidth]{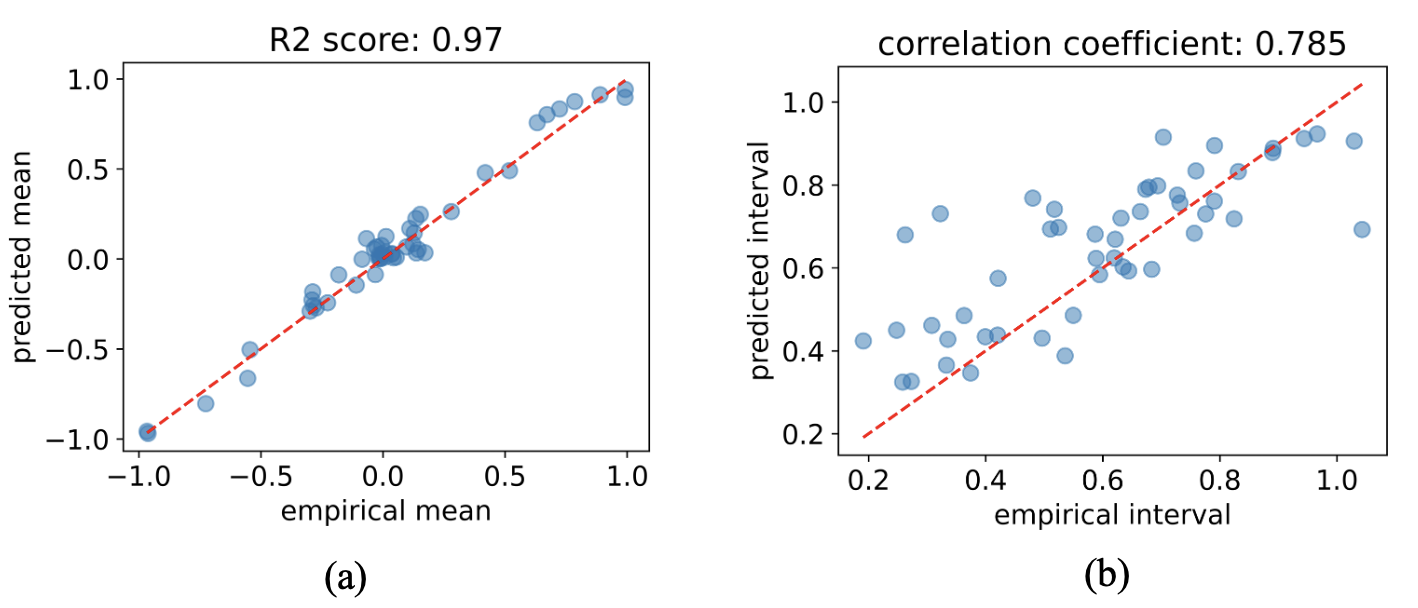}
    \caption{Comparing the predicted mean with the actual or empirical mean in (a), alongside the predicted intervals against the actual intervals in (b) using the optimized PNN on the synthetic data set. The PNN delivers highly accurate predictions of the empirical mean while generating intervals that are marginally wider than those observed empirically.  Note that for both metrics, the highest possible score is $1$.}
    \label{fig:syn_results}
\end{figure}

Figure \ref{fig:syn_results}(a) demonstrates that the first ``head'' of the optimized PNN, illustrated in Figure \ref{fig:pnn}, provides very accurate estimates of the actual mean for all $50$ distinct test data points. Specifically, the R-squared score reaches $0.97$, nearing the ideal score of $1$, and the proximity of the scatter plot to the 45-degree line further attests to the efficacy of the optimized PNN. Furthermore, Figure \ref{fig:syn_results}(b) illustrates the comparison between the predicted and empirical intervals, where we use both the predicted mean and variance produced by the two ``heads'' of the optimized PNN. The slight upward orientation of the data points relative to the 45-degree line suggests a conservative tendency in the PNN, tending to predict slightly broader intervals than observed empirically. However, the Pearson correlation coefficient \cite{zhou2016new} between the actual and predicted interval vectors shows a strong positive correlation, suggesting that despite the slight overestimation, these intervals are still useful in scientific settings for representing aleatoric uncertainty. Note that the correlation coefficient ranges from $-1$ to $1$, and values that are close to 1 signify a strong positive correlation. For instance, as depicted in Figure \ref{fig:syn_visual}(c), the model predicts larger intervals for values of $x$ near 1 compared to those near -1, aligning with the expected behavior from the model specified in Eq.~\eqref{eq:data}.

In the final segment of this section, we delineate a crucial distinction between PNNs with probabilistic output layers and Gaussian process regression (GPR), a widely used technique for quantifying predictive uncertainties in the field of scientific machine learning \cite{schulz2018tutorial,deringer2021gaussian,maddox2022low}. To clarify, we explore the key aspects of GPR, including the model assumption, training methodology, and prediction mechanism. 

A Gaussian process is a collection of random variables, any finite number of which follows a joint Gaussian or normal distribution. Hence, a Gaussian process is fully defined by its mean function, $m(\mathbf{x})$, and its covariance function, $k(\mathbf{x}, \mathbf{x}')$. The covariance function is typically chosen to be the squared exponential kernel function in the form of $k(\mathbf{x}, \mathbf{x}')=\exp\big(-\frac{\|\mathbf{x}-\mathbf{x}'\|_2^2}{2l^2}\big)$, where $l>0$ is an important hyperparameter, known as the bandwidth or length scale \cite{pourkamali2019improved}. A notable benefit of using the scikit-learn package \cite{pedregosa2011scikit} for implementing GPR is its ability to automatically adjust the kernel bandwidth during the fitting process, given specified bounds for the length scale, an approach that we employ in our research.

For a given training data set $\mathcal{D}_{\text{train}} = \{(\mathbf{x}_i, y_i)\}_{i=1}^n$, the GPR observation model is represented as:
\begin{equation}
y_i = f(\mathbf{x}_i) + \varepsilon,\label{eq:gpr}
\end{equation}
where $\varepsilon$ in this case adheres to a zero-mean normal distribution with fixed variance, i.e., $\mathcal{N}(0, \sigma^2)$. Therefore, the joint distribution of the observed outputs in the vector format, that is $\mathbf{y}\in\mathbb{R}^n$, and the function value at a new test point $\mathbf{x}^{(t)}$ is given by:
\begin{equation}
\begin{bmatrix}\mathbf{y} \\ f(\mathbf{x}^{(t)})\end{bmatrix}\sim \mathcal{N}\Big(\mathbf{0}, \begin{bmatrix} \mathbf{K}+\sigma^2\mathbf{I} & \mathbf{k}^{(t)}\\(\mathbf{k}^{(t)})^T & k(\mathbf{x}^{(t)}, \mathbf{x}^{(t)}) \end{bmatrix}\Big),
\end{equation}
where $\mathbf{I}$ is the identity matrix, $\mathbf{K} \in \mathbb{R}^{n \times n}$ is the covariance matrix computed from $\mathcal{D}_{\text{train}}$, and $\mathbf{k}^{(t)} \in \mathbb{R}^{n}$ is a column vector of covariances between the training samples and the test point $\mathbf{x}^{(t)}$. A limitation of standard GPR is its assumption of homoscedastic noise variance, represented by a constant $\sigma^2$. This assumption is less effective in heteroscedastic settings, where the noise level varies, since there is no inherent mechanism to adaptively estimate different noise levels from the data \cite{li2020gaussian}. In the scikit-learn implementation, users have the flexibility to specify $\sigma^2$ as a hyperparameter, allowing them to add a constant value to the diagonal entries of the covariance matrix, thereby adjusting for observation noise through fine-tuning and grid search.

Based on the above model, GPR predictions are made by computing the mean and variance functions as follows \cite{10360364}:
\begin{align}
    f_{\mu}^{\text{GPR}}&= (\mathbf{k}^{(t)})^T (\mathbf{K}+\sigma^2\mathbf{I})^{-1}\mathbf{y}\nonumber \\
    f_{\sigma}^{\text{GPR}}&= k(\mathbf{x}^{(t)}, \mathbf{x}^{(t)}) - (\mathbf{k}^{(t)})^T(\mathbf{K}+\sigma^2\mathbf{I})^{-1}\mathbf{k}^{(t)}. 
\end{align}
Therefore, like PNNs, GPR offers insights into each test input through mean and variance-related outputs. However, it is essential to note that the variance function in GPR mainly reflects the model's epistemic uncertainty, or the uncertainty due to the model's lack of knowledge, rather than the inherent data uncertainty. This differentiation is particularly significant in scientific fields where data or aleatoric uncertainty frequently exhibits heteroscedastic behavior, meaning it varies across different observations. Such variability cannot be adequately represented using a constant noise variance $\sigma^2$, as implied in the standard GPR formulation in Eq.~\eqref{eq:gpr}.

\begin{figure}[ht]
    \centering
    \includegraphics[width=\textwidth]{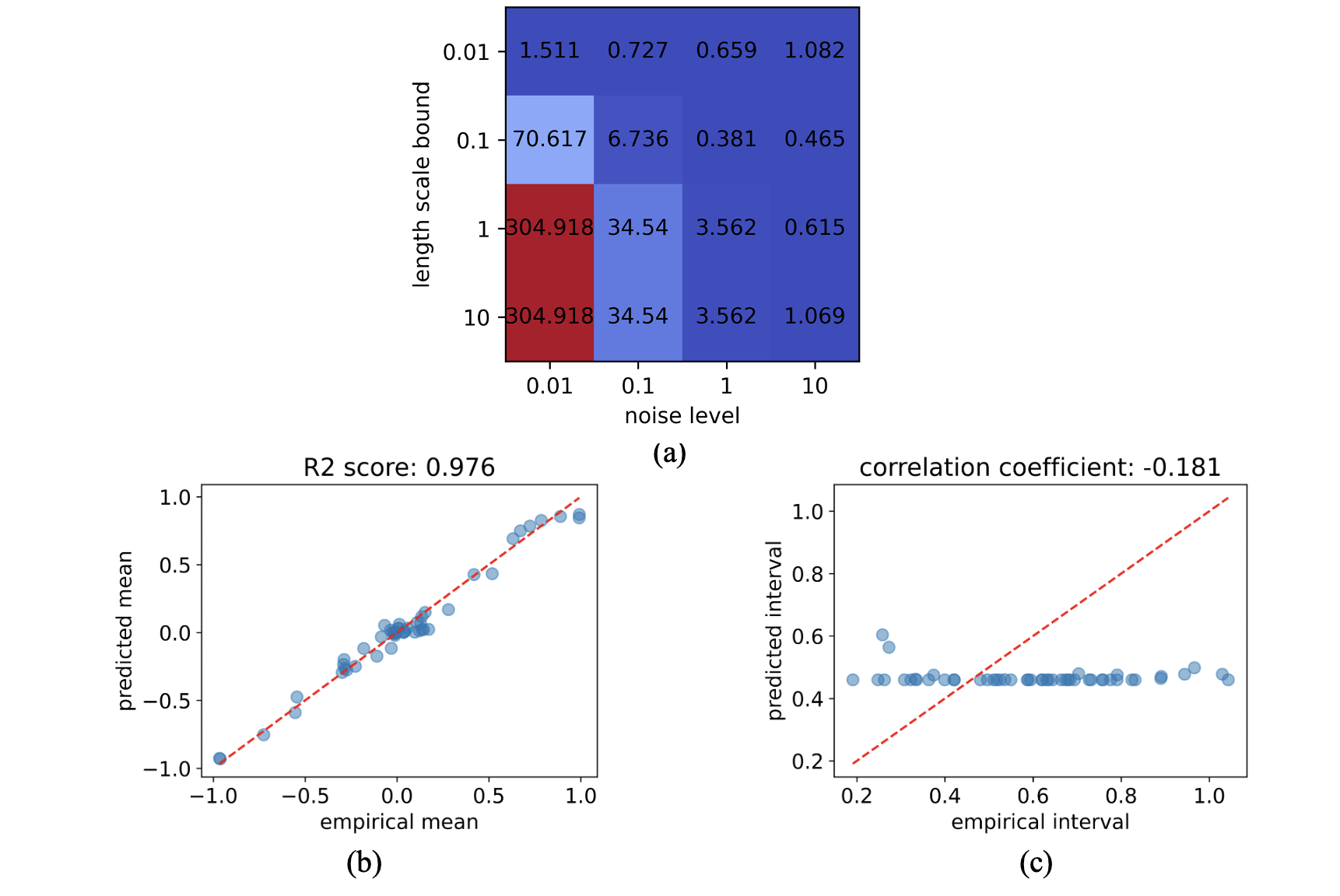}
    \caption{Using GPR to model the heteroscedastic aleatoric noise in the synthetic data set, we display the KL divergence scores for different settings of the two hyperparameters in (a). Additionally, we evaluate the top-performing GPR model's accuracy in predicting mean and variance, as shown in (b) and (c). The results clearly indicate that GPR struggles to adequately represent the aleatoric uncertainty inherent in this controlled data setting.}
    \label{fig:syn_results_gpr}
\end{figure}

To demonstrate this point, we return to the synthetic data set featuring heteroscedastic noise, as described in Eq.~\eqref{eq:data}.  In a manner similar to our approach with PNNs, we begin by fine-tuning two critical hyperparameters for GPR: the upper limit of the kernel's length scale parameter $l$ and the noise variance $\sigma^2$. Following this, we calculate the KL divergence to assess the distance between the predicted normal distribution, defined by the mean $f_{\mu}^{\text{GPR}}$ and variance $f_{\sigma}^{\text{GPR}}$ produced by GPR, and the true normal distribution of the test data set. Figure \ref{fig:syn_results_gpr}(a) illustrates that GPR's effectiveness is highly dependent on the selection of these two hyperparameters. For instance, whereas the highest KL divergence score\textemdash indicating the least similarity between true and predicted distributions\textemdash for PNNs stood at $0.447$, in the case of GPR, this score escalates dramatically to approximately $304$. This discrepancy highlights the increased complexity and resource demand of GPR in identifying suitable hyperparameter settings.
Nonetheless, the optimal GPR configuration, with a length scale limit of $0.1$ and a noise level of $1$, achieves a KL divergence score of $0.381$. This performance is competitive with that of the PNNs, demonstrating that, despite the challenges, GPR can attain comparable accuracy with careful hyperparameter tuning.

Furthermore, we delve into the efficacy of the top-performing GPR model. Figures \ref{fig:syn_results_gpr}(b) and (c) showcase the comparison between the predicted means and empirical means, as well as the predicted intervals versus the actual intervals, for the $50$ distinct input vectors in the test data set. These findings indicate that the finely tuned GPR model yields highly accurate estimates of the empirical means, with an R-squared score of $0.976$. This score slightly exceeds the best performance achieved by PNNs, which was $0.97$. However, a significant limitation of the GPR model becomes apparent in Figure \ref{fig:syn_results_gpr}(c). As discussed previously, standard GPR assumes constant observation noise (homoscedasticity), rendering it incapable of adequately capturing aleatoric noise, even within our synthetic data set. Notably, the correlation coefficient is slightly below $0$, implying a lack of meaningful correlation between the predicted and actual intervals. This undermines the primary advantage of employing GPR to quantify uncertainties in this scenario. Consequently, our comparative study, coupled with the algorithmic review, demonstrates that PNNs exhibit a superior capacity for modeling heteroscedastic aleatoric or data uncertainty.

\section{Benchmarking with the Ishigami Function}\label{sec:exp-syn}
In this section, we employ the Ishigami function, as referenced in \cite{allaire2012variance,opgenoord2016variance,hariri2022structural}, to demonstrate the efficacy of PNNs to model the aleatoric uncertainty for multidimensional inputs. The Ishigami function, a well-regarded benchmark in the field, is defined for a 3-dimensional input vector $\mathbf{x} = [x_1, x_2, x_3]$ as follows:
\begin{equation}
f(\mathbf{x}) = \sin(x_1) + a\sin^2(x_2) + bx_3^4\sin(x_1),\label{eq:ishigami}
\end{equation}
where $x_1, x_2, x_3$ are independent variables uniformly distributed over the range $[-\pi, \pi]$, and $a$ and $b$ are constants; we use $a=7$ and $b=0.1$ in our study. The Ishigami function is particularly challenging and thus a robust benchmark for two main reasons: (1) it incorporates nonlinear elements such as sine and polynomial terms, and (2) it is nonmonotonic with its variables, meaning that its output does not consistently increase or decrease with changes in $x_1, x_2,$ or $x_3$. These characteristics introduce significant complexity, making the Ishigami function an ideal testbed for assessing the modeling capabilities of PNNs. Additionally, to explore aleatoric uncertainty, we examine the following relationship between input vectors $\mathbf{x}$ and their corresponding outputs $y$:
\begin{equation}
    y = f(\mathbf{x}) + \varepsilon, \;\;\varepsilon\sim \mathcal{N}(0, 0.2\;|f(\mathbf{x})|).
\end{equation}
Hence, this variance structure introduces heteroscedastic noise, making the uncertainty in the output $y$ dependent on the magnitude of $f(\mathbf{x})$, which is defined in Eq.~\eqref{eq:ishigami}. For the training phase, we create $10$ instances of the noise term $\varepsilon$ for each of the $300$ training inputs, yielding a total of $3,\!000$ training pairs. In a similar vein, for the evaluation phase, we generate $10$ noise realizations for each of the $100$ test data points. All these data points are selected uniformly at random from the entire input space $[-\pi, \pi]^3$. Also, consistent with the approach in the prior case study, we set the batch size at $32$ and the total number of training epochs at $100$.

\begin{figure}[ht]
    \centering
    \includegraphics[width=\textwidth]{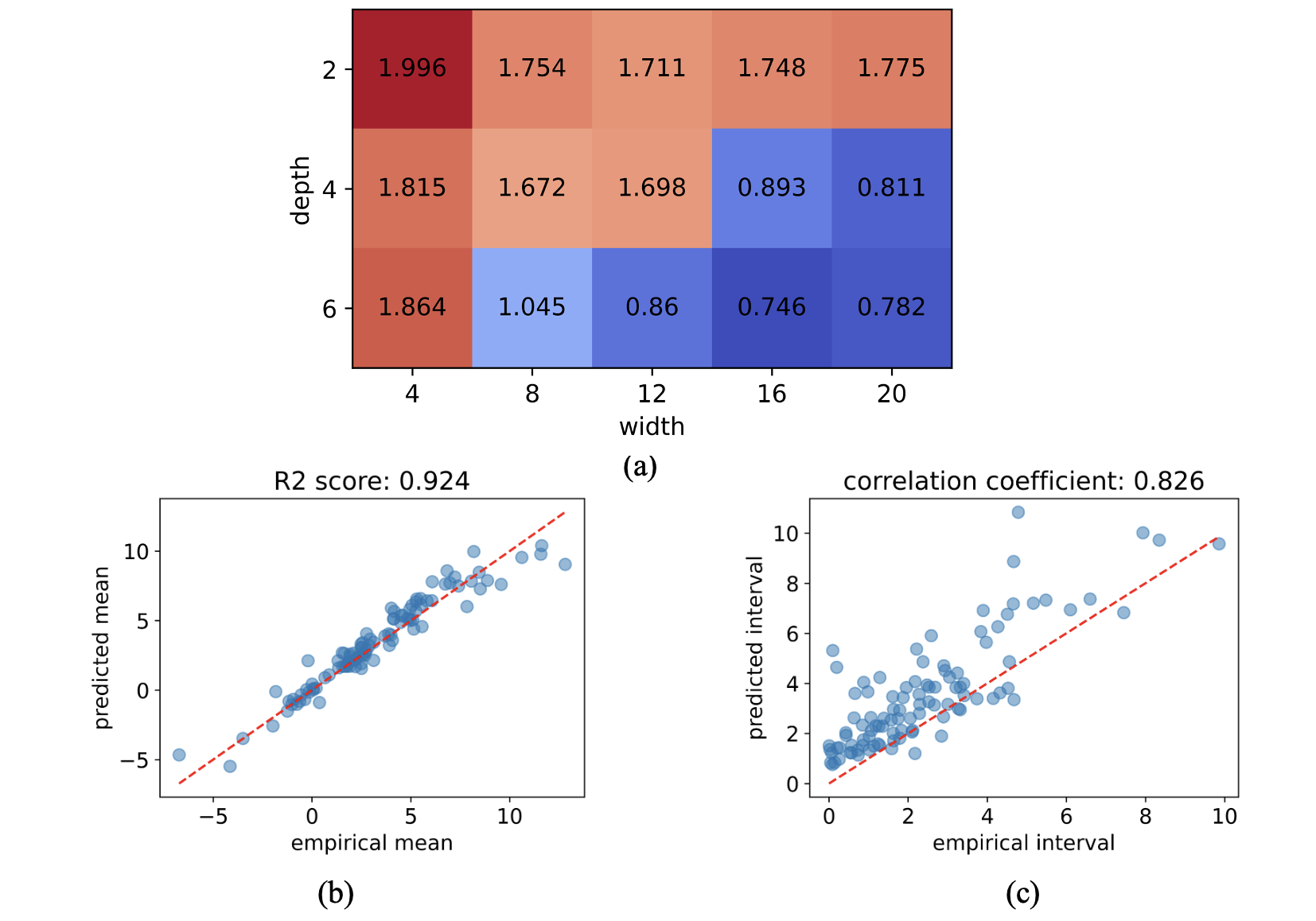}
    \caption{Utilizing PNNs to model the aleatoric uncertainty associated with the Ishigami function, we report the KL divergence values between the actual and estimated output distributions in (a). Additionally, we examine the predicted mean and interval accuracies in (b) and (c). The results indicate that the optimized PNN is effective in generating an accurate predictive distribution for the output.}
    \label{fig:ishigami}
\end{figure}

Figure \ref{fig:ishigami}(a) presents the outcomes of a grid search exploring three different depths (number of hidden layers) $\{2,4,6\}$ and five widths (number of hidden units) $\{4,8,12,16,20\}$. This search was motivated by the nonlinear nature of the input-output relationship in the case study. The findings reveal that PNNs with $2$ hidden layers fail to capture the output distributions accurately, even with a substantial width of $20$ units. Conversely, enhancing the network's depth significantly improves prediction accuracy. Notably, a depth of $6$ layers with various widths, such as $16$ and $20$ hidden units, yields considerably lower KL divergence values. Within the tested hyperparameter range, the optimal configuration for the PNN was found to be a depth of $6$ layers and a width of $16$ units, which achieved the minimal KL divergence score.

To delve deeper into the predictive capabilities of the optimized PNN, we display a comparison of the predicted mean $f_{\mu}(\mathbf{x}^{(t)};\boldsymbol{\theta}^*)$ against the empirical mean $f_{\mu}^{\text{emp}}(\mathbf{x}^{(t)})$ for each test vector $\mathbf{x}^{(t)}$ in Figure \ref{fig:ishigami}(b). This comparison illustrates that the predicted means align closely with the actual values, clustering tightly near the 45-degree line. Furthermore, the achieved R-squared score  is $0.924$, which is close to the ideal score of $1$. Although this R-squared score is marginally lower than what was observed in the preceding 1D example in Figure \ref{fig:syn_results}(a), it is important to acknowledge the increased complexity introduced by the Ishigami function's greater nonlinearity.

We next turn our attention to the prediction intervals generated by the optimized PNN, which incorporates both the predicted mean $f_{\mu}(\mathbf{x}^{(t)};\boldsymbol{\theta}^*)$ and variance $f_{\sigma}(\mathbf{x}^{(t)};\boldsymbol{\theta}^*)$ to construct a 90\% confidence interval. Figure \ref{fig:ishigami}(c) reveals a mild upward trend in the distribution of the scatter plot points, indicating that the optimized PNN tends to slightly overpredict the width of the true prediction intervals, particularly for larger intervals. Despite this, the strong positive correlation between the estimated and actual intervals, as evidenced by a correlation coefficient of $0.826$, underscores the model's reliability in capturing the relationship. This performance even surpasses that observed in the 1D example shown in Figure \ref{fig:syn_results}(b). Consequently, we deduce that PNNs are capable of delivering precise and reliable representations of aleatoric uncertainty within multidimensional and nonlinear systems in scientific problems.

\section{Real-world Application in Material Science: Composite Microstructure Generation}\label{sec:exp-real}
Fiber-reinforced composites are highly sought after in engineering applications that demand high strength with minimal weight. For these composites to be effectively used in structural parts, it is crucial to develop computer models that can accurately predict their behavior under various loads. The challenge in modeling composites lies in the need for high fidelity to capture microscale details, such as densely packed fiber clusters or resin-rich zones, which can significantly increase computational demands. To address this, multiscale modeling has proven effective, allowing for the detailed representation of microscale features within smaller, more manageable models. 

In this work, we used a microstructure generator that employs several input parameters to shape the fiber morphology, maintaining a balance between controlled outcomes and inherent randomness \cite{husseini2023generation,husseini2023generation2}.  The generator was developed using in-house code with the Discrete Element Method as its foundation.   Fiber seeding parameters control the size of the microstructure and the initial placement or seeding of the fibers prior to enforcing contact between fibers. First, a number of fibers, $n_f$, were placed in a bounded region divided into cells that were padded by a margin, $m_{cell}$, based on a user-defined number of fibers per cell, $n_{f/cell}$, until a global volume fraction, $V_f$, was reached.  Increasing the margin decreased the area where fibers could be placed, increasing the chance of initial overlapping and the  potential energy of the system. Once seeded, the fibers were allowed to disperse, with movement initiated by contact between the fibers.

The simulation phase of this generator continued until a prescribed kinetic energy cutoff, $\epsilon_{KE}$, criteria was met.  Initially, a prescribed number of relaxation iterations was run without damping until a maximum number of time steps, $t_{n_{max}}$, was reached and damping was enforced. Three different types of damping were used to alter the kinetic energy dispersion: contact damping, $C_{ij}^{n}$, global damping, $C_i$,  which increased with each time step, and incremental damping, $d_i$, which was enforced at each time step. Once the simulation ended, a minimum spacing between the fibers, $d_{min}$, was created by reducing the radius of each fiber. 

\begin{figure}[ht]
    \centering
    \includegraphics[width=\textwidth]{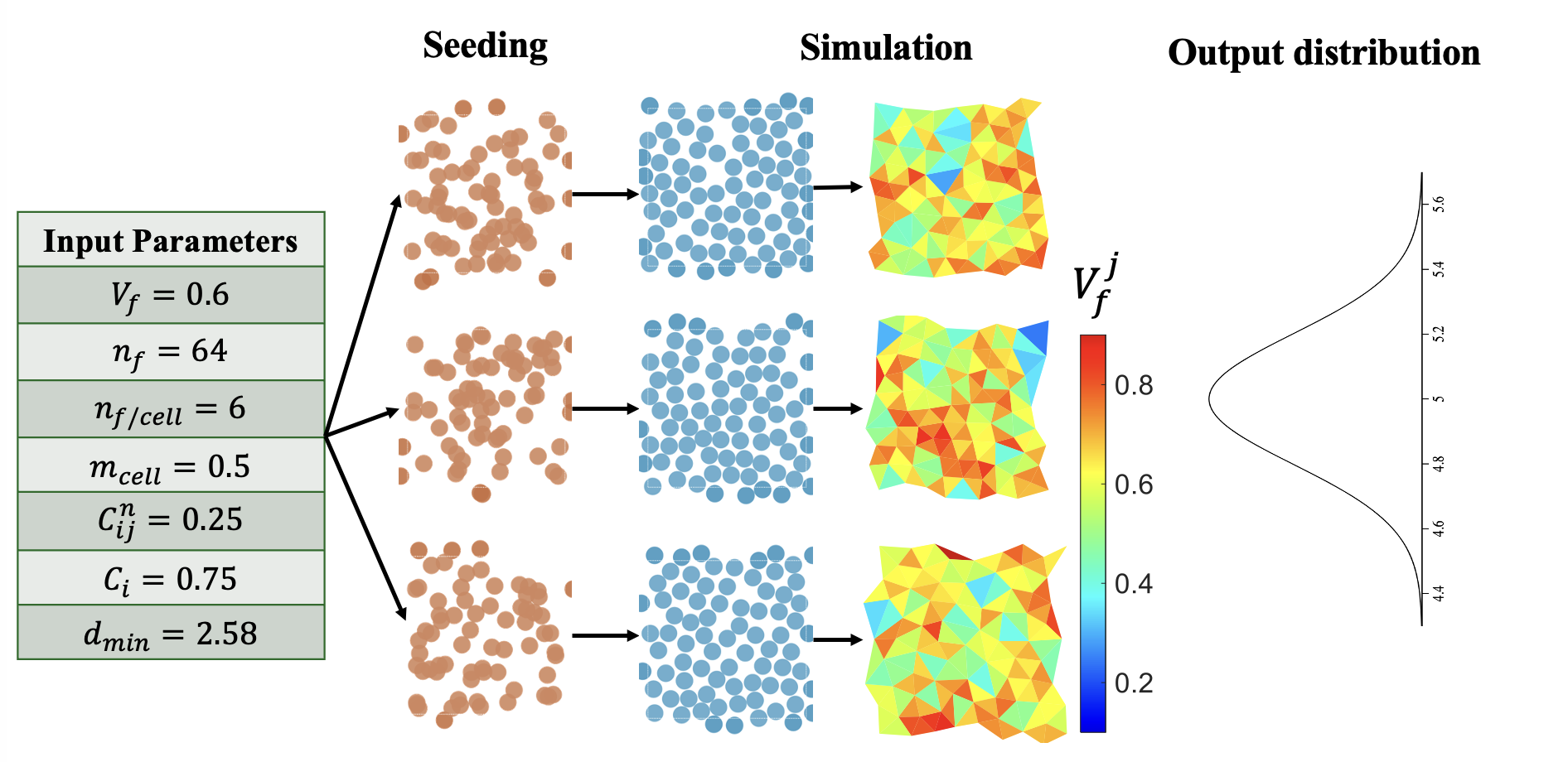}
    \caption{Aleatoric uncertainty in generating microstructures: the observation model for any given input vector $\mathbf{x}\in\mathbb{R}^7$ incorporates a heteroscedastic noise term, which varies with the input. Consequently, a data set comprising nearly $50$ samples of the (unknown) noise term for each input vector has been constructed, enabling the modeling of aleatoric uncertainty through PNNs.}
    \label{fig:inputcombos}
\end{figure}

In this case study, a given set of inputs for the microstructure generator does not ensure a consistent fiber morphology and the resulting descriptors, due to the initial random placement of fibers, as shown in Figure \ref{fig:inputcombos}. This randomness originates from the initial setup, where fibers are positioned randomly, potentially leading to overlaps. As the simulation begins, the movements and interactions of the fibers are dictated by their initial placements, leading them to disperse and collide until a stable state is achieved through contact dynamics and viscous damping. 

In summary, the stochastic nature of this process leads to diverse outcomes in fiber configurations across simulations, reflected in a spectrum of microstructural descriptor values for each set of input parameters. This variability introduces a heteroscedastic noise element into the relationship between the input vector $\mathbf{x}$ and the output variable $y$. To address this, we performed around 50 repetitions for each set of inputs, generating 50 distinct versions of the noise term. In this particular study, we have a total of $23,\!465$ unique input vectors. We allocate 20\% of these inputs to the test data set, with the remaining 80\% used for training purposes. Considering that the training data set encompasses roughly one million input-output pairs, we adjust the batch size to $2,\!048$ and reduce the number of epochs to $10$.

\begin{figure}[ht]
    \centering
    \includegraphics[width=0.5\textwidth]{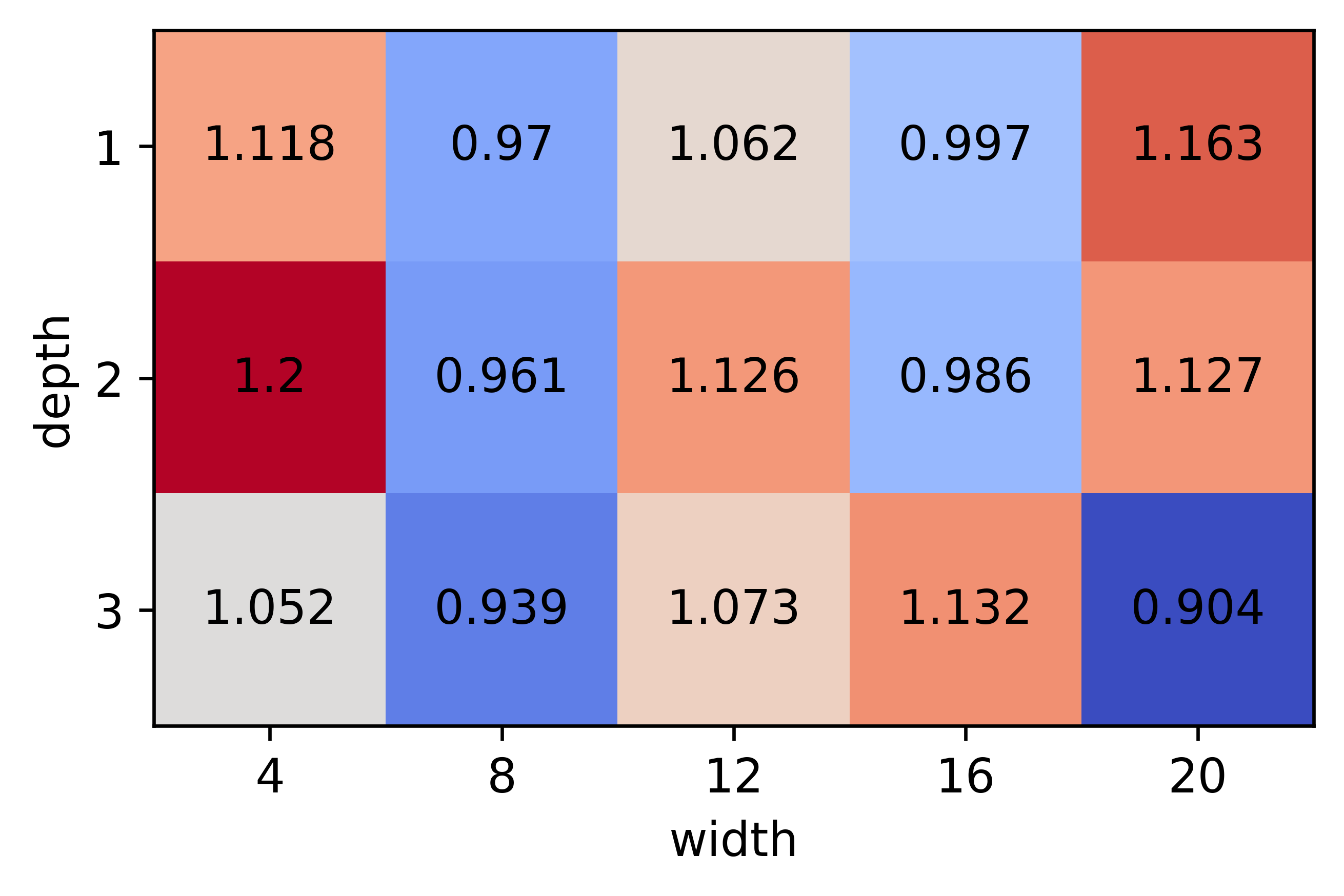}
    \caption{Presenting the KL divergence scores achieved by PNNs across different depths and widths in the context of the microstructure generation problem. Our case study highlights multiple architectures achieving KL divergence scores under 1.}
    \label{fig:cm-nasa}
\end{figure}

Figure \ref{fig:cm-nasa} presents the outcomes of a grid search conducted on PNNs with various configurations of depth and width. This study reveals that even a single hidden layer can yield relatively low KL divergence scores, with the scores dropping below $1$ for configurations where the network width, or the number of hidden units, is set to $8$. Notably, a width of $8$ consistently results in KL divergence values under $1$ across all tested depths, highlighting an optimal width. Consequently, we will further explore the performance of PNNs with a fixed width of $8$ and depths ranging from 1 to 3. However, it is important to highlight that the lowest KL divergence observed was with a configuration of 3 hidden layers, each comprising 20 units. We will next elaborate on how these configurations influence the predicted output distribution.

\begin{figure}[ht]
    \centering
    \includegraphics[width=\textwidth]{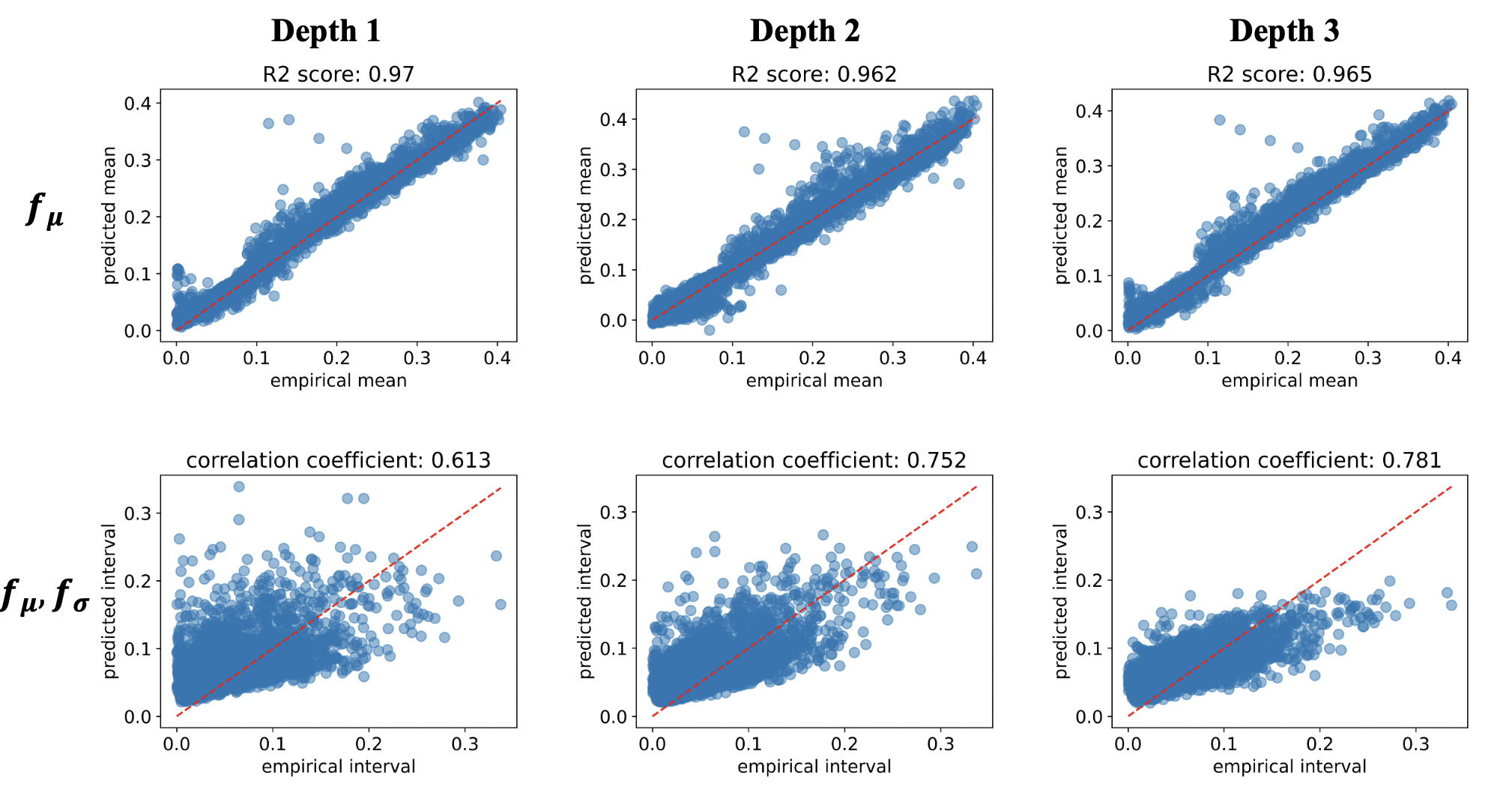}
    \caption{This figure displays the comparison of predicted versus empirical means in the top row, and the comparison of predicted versus empirical intervals as found in our test set in the bottom row. Throughout these comparisons using the microstructure generation problem, each PNN maintains a fixed width of $8$, with the depth adjusted from $1$ to $3$.}
    \label{fig:width}
\end{figure}

Based on the results obtained from our grid search, we further explore the predictive performance of PNNs with a consistent width of 8 across varying depths, as illustrated in Figure \ref{fig:width}. It is important to note that the KL divergence scores showed a consistent decrease with increasing depth, indicating a more accurate representation of the true or empirical output distribution. This trend aligns with the observations in Figure \ref{fig:width}, especially when comparing the predicted intervals with the empirical intervals. For instance, at a depth of 1, the correlation coefficient stands at 0.613, but this value escalates to 0.781 with the depth increased to 3. Moreover, with a depth of 3 and width of 8, the data points demonstrate a closer alignment to the 45-degree line compared to other configurations. From these observations, we draw two primary conclusions: firstly, the performance of PNNs exhibits a robustness to variations in network architecture, which is advantageous from a practical perspective. Secondly, PNNs are capable of accurately estimating aleatoric uncertainty within this materials science application.

\begin{figure}[ht]
    \centering
    \includegraphics[width=\textwidth]{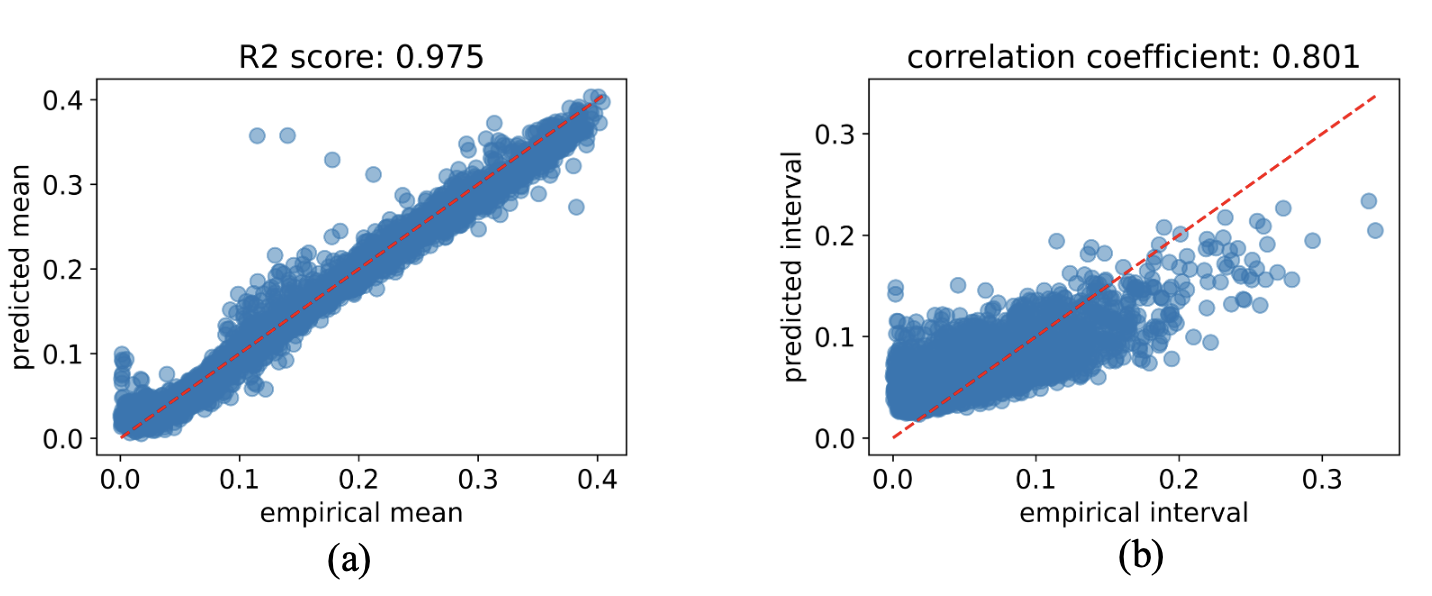}
    \caption{Assessing the effectiveness of the PNN model that achieved the minimal KL divergence score in our grid search. The findings demonstrate the model's capability to deliver dependable estimations of both the mean of the output data and the heteroscedastic aleatoric noise within the context of a real-world scientific machine learning problem.}
    \label{fig:nasa_best}
\end{figure}

In the final segment of this section, we delve into the performance of the optimized PNN, which recorded the lowest KL divergence score during our grid search. Figure \ref{fig:nasa_best} presents a comparison between predicted and empirical means, alongside predicted and empirical intervals, for the refined PNN that includes $3$ hidden layers, each consisting of $20$ units. The R-squared score depicted in Figure \ref{fig:nasa_best}(a) indicates a modest improvement in the predicted mean, especially when compared to the R-squared scores shown in Figure \ref{fig:width}, where the network width was fixed at 8. This reinforces the effectiveness of PNNs in approximating the mean of the underlying normal distribution that generates the observed data. Moreover, the correlation coefficient for the predicted intervals shows a slight enhancement, being approximately $0.02$ higher than the previously highest reported value. Consequently, this underscores the utility of the KL divergence score as a valuable metric for identifying precise PNN configurations in the context of real-world scientific machine learning challenges.

\section{Conclusion}\label{sec:conc}
In this paper, we explored various facets of developing probabilistic neural networks (PNNs) to model aleatoric uncertainty, focusing on aspects such as model assumptions, loss functions, and employing Kullback-Leibler (KL) divergence to assess the accuracy of predictions by comparing the predicted and actual output distributions. Utilizing KL divergence to steer our grid search yielded several notable findings. First, PNN performance appears relatively robust to network architecture variations, such as depth and width. For instance, in a practical material science application, we found that different PNN configurations could all generate satisfactory predictive models. Second, PNNs demonstrated the capability to accurately estimate the empirical mean under a range of simulated or controlled aleatoric noise conditions, as well as in a real-world scenario with an indeterminate noise analytical form. Third, the prediction intervals provided by PNNs closely matched the empirical prediction intervals, despite occasional slight overestimations. However, the strong correlation coefficient underscores the utility of these intervals for decision making, particularly in predicting the consistency of real-world system outputs from repeated inputs, which is crucial for optimal design strategies. Although this study primarily addresses aleatoric uncertainty due to its pressing significance, future work will aim to enhance PNNs for concurrent modeling of aleatoric and epistemic uncertainties, paving the way for advanced active learning methodologies.

\section*{Acknowledgements}
The present work was partially supported by a NASA Space Technology Graduate Research Opportunity (80NSSC21K1285) and a NASA NRA (80NSSC21N0102) through the NASA Transformational Tools and Technologies (TTT) program, under the Aeronautic Research Mission Directorate (ARMD). 
This work was completed in part with resources provided by the University of Massachusetts' Green High Performance Computing Cluster (GHPCC).

\section*{Conflicts of Interest}
The authors declare that they have no known competing financial interests or personal relationships that could have appeared to influence the work reported in this paper.


 \bibliographystyle{elsarticle-num} 
 \bibliography{sample.bib}





\end{document}